\documentclass[times,twocolumn,final,authoryear]{elsarticle}
\usepackage[a4paper, top=3.2cm, bottom=4cm, left=1.5cm, right=1.5cm]%
{geometry}
\usepackage{framed,multirow}

\usepackage{amsmath}

\usepackage{amssymb}
\usepackage{latexsym}

\usepackage{url}
\usepackage{xcolor}
\definecolor{newcolor}{rgb}{.8,.349,.1}
\usepackage{subfigure}
\newtheorem{theorem}{Theorem}
\usepackage{mathtools}
\usepackage{multirow}
\usepackage{multicol}
\usepackage{booktabs}

\begin{document}


\title{Dual affine moment invariants}
\author{You Hao, Hanlin Mo, Qi Li, He Zhang, Hua Li}

%
%

\begin{abstract}
Affine transformation is one of the most common transformations in nature, which is an important issue in the field of computer vision and shape analysis. And affine transformations often occur in both shape and color space simultaneously, which can be termed as \textit{Dual-Affine Transformation} (DAT). In general, we should derive invariants of different data formats separately, such as 2D color images, 3D color objects, or even higher-dimensional data. To the best of our knowledge, there is no general framework to derive invariants for all of these data formats. In this paper, we propose a general framework to derive moment invariants under DAT for objects in $M$-dimensional space with $N$ channels, which can be called dual-affine moment invariants (DAMI). Following this framework, we present the generating formula of DAMI under DAT for 3D color objects. Then, we instantiated a complete set of DAMI for 3D color objects with orders and degrees no greater than 4. Finally, we analyze the characteristic of these DAMI and conduct classification experiments to evaluate the stability and discriminability of them. The results prove that DAMI are robust for DAT. Our derivation framework can be applied to data in any dimension with any number of channels.
\end{abstract}

%



\maketitle

\section{Introduction}
The world we live in is so rich and varied. But there must be some invariable factors that enable people to identify the same object from the change. 
Among these complex transformations in nature, affine transformation is one of the most basic and common issue in computer vision, which is the largest set of linear transformations, including rotation, translation, scaling, et cetera. 
Geometric affine transformation in shape space can be used to approximate the projective transformations in images where the camera happened a slight change of viewing perspective, which is the most common situation in the imaging process. 
Photometric affine transformation in channel space is the best linear model to describe the color degradation of the outdoor scenes, which has been proved in \cite{geusebroek2001color}. 
And affine transformations often occur in both shape and color space simultaneously, which can be termed as \textit{Dual-Affine Transformation} (DAT). DAT is a useful and common transformation model to describe the degradations happened in vision information from our diverse and colorful world, which considering both shape and color degradation.
Invariants under DAT is a kind of important image feature that can be used for object classification and object retrieval.

With the diversification of visual data acquisition means, the formats of visual data are also diversified, which exist in different dimension and with different number of color channels, such as 2D color images, 3D color objects, 4D movies of a 3D color object, even higher-dimensional data and multi-spectrum images with multi-channels. Other data organized in channels (such as vector fields) can also be included in this work. Some examples are shown in Figure \ref{fig:data}.


In common, we can derive dual-affine moment invariants for objects in a particular dimension with a certain number of channels separately. To the best of our knowledge, there is no general framework to derive invariants for all of these data formats. In this paper, 
we systematically analyze the properties of invariants under DAT for different data dimensions and present a general framework for deriving invariants under DAT for objects in $M$-dimensional shape space with $N$ color channels.

\subsection{Object definition in $M$-dimensional space with $N$ channels}

An object in $M$-dimensional ($M$-D) shape space, ($X,Y,Z,S4,S5,...$), with $N$ channels, ($R,G,B,C4,C5,...$), can be described as Equation (\ref{equ:fun}):
\begin{align}\label{equ:fun}
O_{M,N} = f_{M,N}(\vec{X}) = (&R(\vec{X}); G(\vec{X});B(\vec{X}); C4(\vec{X});C5(\vec{X});...)
\end{align}	
where $\vec{X}$ represents the coordinate $(x,y,z,s4,s5,...)^T$ in $M$-D space. 
And $R(\cdot), G(\cdot), B(\cdot), C4(\cdot), C5(\cdot)$ represent the functions in different channels.
Object $O$ can be defined in another form as Equation (\ref{equ:obj}):
\begin{align}\label{equ:obj}
O_{M,N}(id) = &[S_{M}(id),C_N(id)] \notag \\
=&[(x_{id},y_{id},z_{id},s4_{id},s5_{id},...)^T,(r_{id},g_{id},b_{id},c4_{id},c5_{id}...)^T] 
\end{align}
where $id$ represents the index of pixels of object $O$. $S_M$ represents the coordinate in shape space and $C_N$ represents the values in channel space.

\begin{figure}
	\centering
	\subfigure[]{
		\includegraphics[width=0.18\linewidth]{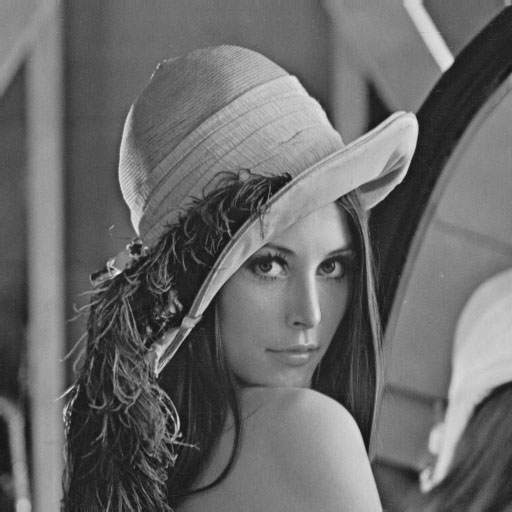}}
	\subfigure[]{
		\includegraphics[width=0.18\linewidth]{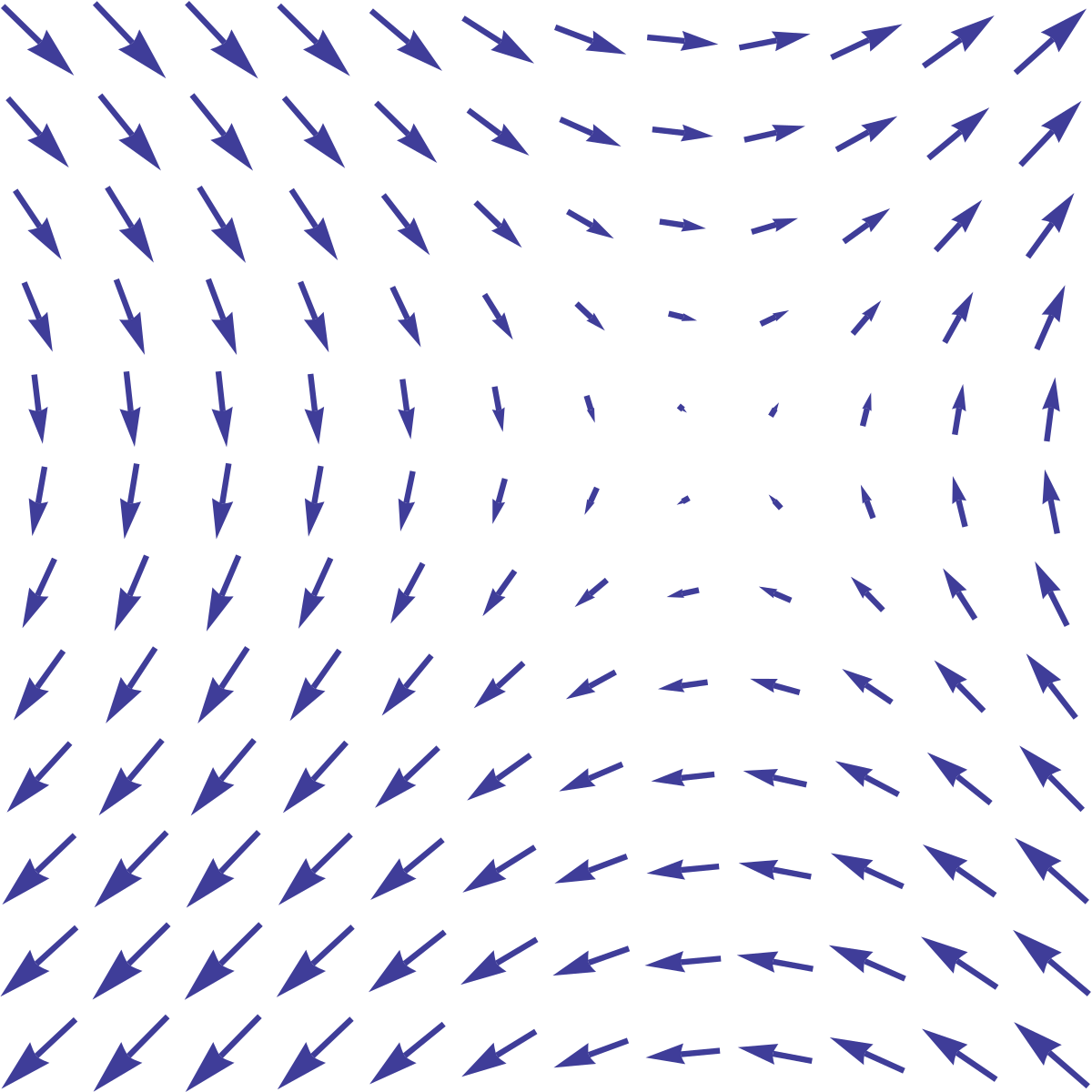}}
	\subfigure[]{
		\includegraphics[width=0.18\linewidth]{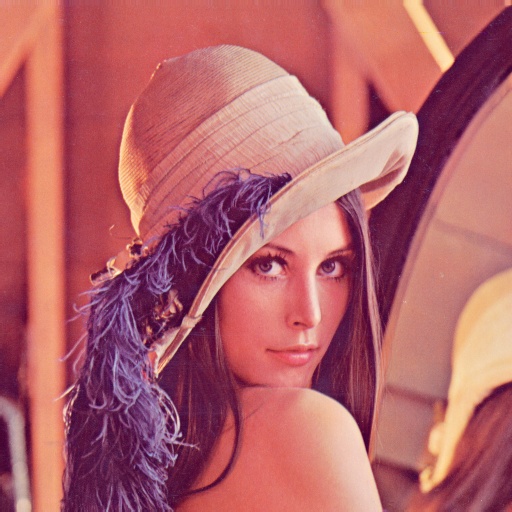}}
	\subfigure[]{
		\includegraphics[height=0.065\textheight]{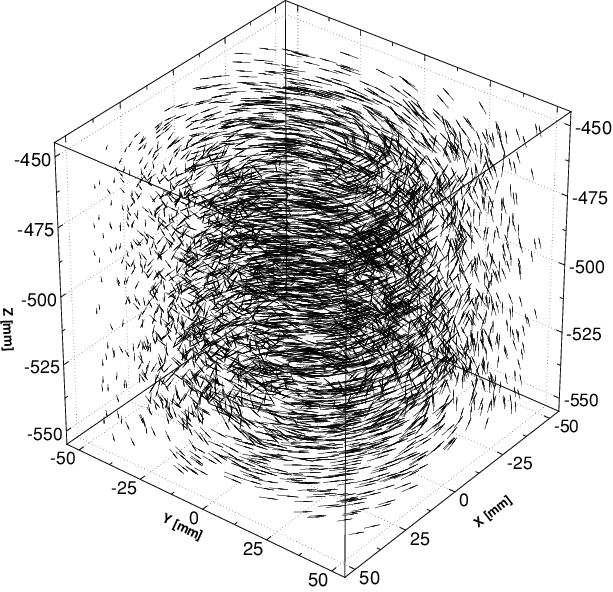}}	
	\subfigure[]{
		\includegraphics[height=0.065\textheight]{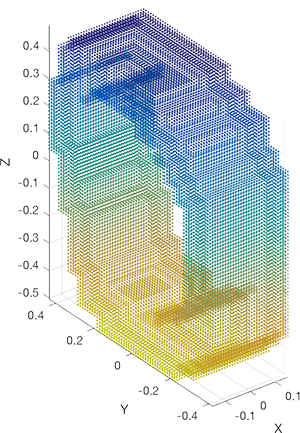}}
	\caption{Examples of different data formats: (a) Gray image in 2D space with 1 channel ($M=2, N=1$); (b) Vector field in 2D space ($M=2, N=2$); (c) Color image in 2D space with 3 channels ($M=2, N=3$); (d) Vector field in 3D space ($M=3, N=3$); (e) Color object in Point Cloud in 3D space with 3 channels ($M=3, N=3$).}
	\label{fig:data}	
\end{figure}

Lots of data are organized in channel forms, which can represent different meanings in different data, such colors in color objects and vectors in vector fields. 
When we specify $M, N$ to different values, we can get different data formats.
For examples, gray image is object in 2D space with 1 channel where $M=2, N=1$.
2D vector field can be regarded as object in 2D space with 2 channels where $M=2, N=2$. The values of 2 channels represent the vector at each point.
Color image is object in 2D space with 3 channels where $M=2, N=3$. 
Both 3D vector field and 3D Point Cloud are objects in 3D space with 3 channels. The difference is that values of 3 channels in 3D vector field represent the vector at each point and that in 3D Point Cloud represent the color at each point.
Some objects in different formats are in Figure \ref{fig:data}.

\subsection{Dual-Affine Transformation (DAT)}

The definition of $M$-D Affine Transformation Matrix (ATM) is listed in Equation (\ref{equ:ts}), which is a $M\times M$ matrix:

\begin{equation} \label{equ:ts}
ATM_M = \left( 
\begin{array}{ccccc}
a_{11} & a_{12} & a_{13} & ... & a_{1M}\\
a_{21} & a_{22} & a_{23} & ... & a_{2M}\\
a_{31} & a_{32} & a_{33} & ... & a_{3M}\\
\vdots & \vdots & \vdots & \ddots & \vdots \\
a_{M1} & a_{M2} & a_{M3} & ... & a_{MM}\\
\end{array}		
\right)_{M\times M}
\end{equation}

After $M$-D affine transformation, the coordinates of object pixels in space $S_M=(x,y,z,s4,s5,...,sM)^T$ should be transformed to  $S^*_M=(x^*,y^*,z^*, s4^*, s5^*,...sM^*)^T$, which can be expressed by Equation (\ref{equ:tsxyz}):
\begin{equation}\label{equ:tsxyz}
S^*_M = \left(\begin{array}{c}
x^* \\ y^* \\ z^* \\ s4^* \\ s5^* \\ ... \\ sM^*
\end{array}\right)
=ATM_M \cdot 
\left(\begin{array}{c}
x \\ y \\ z \\ s4 \\ s5 \\ ... \\ sM
\end{array}\right) = ATM_M \cdot S_M
\end{equation}

After $N$-D affine transformation, the pixel values at different channels $C_N=(r,g,b,c4,c5,...)^T$ will be transformed to $C^*_N=(r^*,g^*,b^*,c4^*,c5^*,...)^T$ in the same way as Equation (\ref{equ:tcrgb}):

\begin{equation}\label{equ:tcrgb}
C^*_N = ATM_N \cdot C_N
\end{equation}

\begin{figure}[t]
	\centering
	\includegraphics[width=\linewidth]{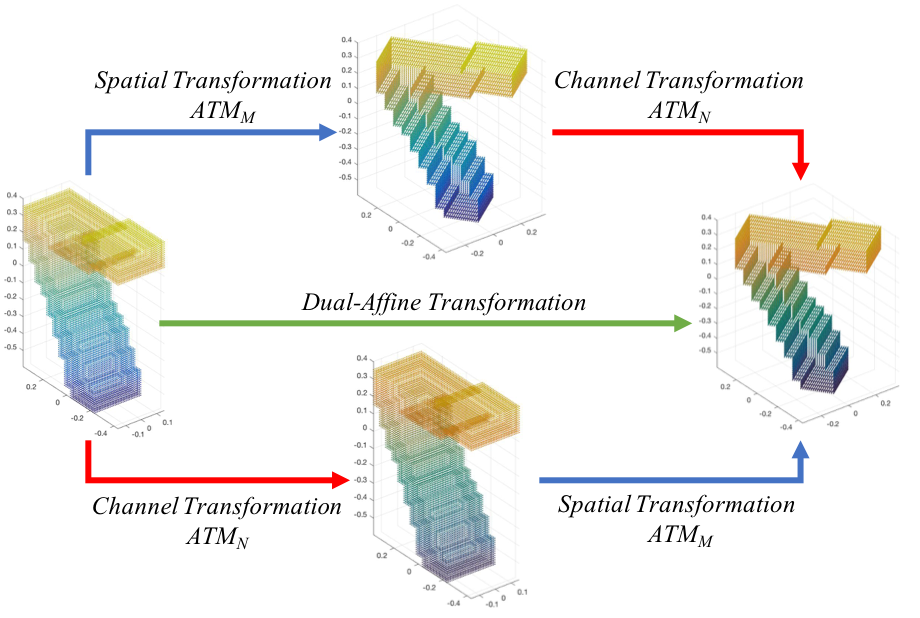}
	\caption{The relationship between \textit{Spatial Transformation}, \textit{Channel Transformation} and DAT with an example of a 3D color Point Cloud ($M=3,N=3$). }
	\label{fig:dat}
\end{figure}

Affine transformations in shape space and channel space often occur simultaneously, which can be called \textit{Dual-Affine Transformation} (DAT). DAT is the transformation we focus on in this article.
The formal definition of DAT for $M$-D objects with $N$ channels can be described as Equation (\ref{equ:dual}):
\begin{equation} \label{equ:dual}
f^*_{M,N}(\vec{X}) = ATM_N \cdot f_{M,N}(ATM_M^{-1} \cdot \vec{X})
\end{equation}
where $ATM_M$ is the \textit{Spatial Transformation} and $ATM_N$ is the \textit{Channel Transformation}. We depict the relationship between them with DAT in Figure \ref{fig:dat}.
DAT is equivalent to applying a \textit{Spatial Transformation} and a \textit{Channel Transformation} on the object independently.

One typical case of DAT is for color images, where $M=2, N=3$. The 2D geometric affine transformation in space represents the deformation of object's shape in the images, and the 3D photometric affine transformation in channel space represents the deformation of object's color. Moment invariants under DAT of color images has been fully discussed in \cite{gong2017naturally}.
Another case is invariants under DAT in 2D vector fields, which have been presented in \cite{schlemmer2007moment,kostkova2018affine,j2019affine}. There are also researches about 3D moment invariants under single affine transformation for 3D objects with one channel in \cite{lo19893,xu2008geometric}.
However, with the continuous advancement of data acquisition methods, more and more forms of data have emerged, like 3D color objects, 4D or higher-dimensional data, multi-spectral images with more than 3 channels, etc. Invariant descriptors under dual-affine transformations in these domains are highly desirable. 

The main contribution of this paper is that, we present a general framework for the derivation of moment invariants under DAT for $M$-dimensional objects with $N$ channels. 
And with this framework, we get a set of independent and usable DAMI for 3D color objects with 3 channels.
The paper is organized as follows. First, we give a survey of relevant literature about DAMI in Section 2. In Section 3, we illuminate the DAMI's derived framework in detail and get the DAMI generation formula for objects in any data format.
In Section 4,  we instantiate DAMI with data format of 3D color point cloud as an example. 
Through a visual approach, we generate a complete set of DAMI equations with $M=3, N=3$.
We perform independent analysis on all DAMI equations and obtain a set of independent DAMI.
In Section 5, we evaluate the ability of DAMI through numerical experiments on several data sets.


\section{Related Works}
Moment invariants were introduced into visual pattern recognition by \cite{hu1962visual} for character recognition, which are invariant under 2D \textit{similarity transformation}.
These seven invariants are considered as a kind of shape descriptor, which plays an important role in the field of pattern recognition. Since then, various kinds of moment invariants have been proposed for different data formats. 
Researchers in this field mainly focus on the following three directions: (i) The first is to extend the transform group of invariants to more complex ones. (ii) The second is to derive invariants for data of different dimensions or formats. (iii) The third is to derive invariants including both shape and color degradations into consideration.

For gray-scale images, Flusser and Suk extended Hu's moment invariants into \textit{affine transformation} in \cite{flusser1993pattern}, which are more common in nature than similar transformation.
Relevant works are also demonstrated in \cite{flusser1994affine,suk2004graph,suk2011affine}. 
Invariants under \textit{projective transformation}, which is the exact transformation model of the imaging process, has been analyzed in \cite{weiss1988projective,suk2004projective,li2018image}.
There are also some other interesting transformations that not only for image data. Invariants under these transformations are also an important direction in the field, like chiral invariants in \cite{zhang2017fast}, invariants under \textit{M\"obius transformation} in \cite{zhang2018differential}, and even some researches about invariants under \textit{conformal}, \textit{quasi-conformal}, and \textit{diffeomorphism transformations} in \cite{white1973global,willmore2000surfaces,zeng2011registration,rustamov2013map}.
Invariant primitive is the most important part of the derivation process of invariants and is the essential invariant property under a certain transform group. Invariant construction frameworks are always based on primitives.
Under affine transformation in 2D space, the cross-product of the coordinates of two points is the covariant primitive, which is used to derive affine invariant for gray-scale images in \cite{suk2004graph}. In \cite{xu2008geometric}, the distances, areas and volumes of coordinates of points are considered as the covariant primitives for different transformations. The essential primitive under affine transformation in $n$-D space is considered as the determination of the coordinates of $n$ pixels in \cite{mamistvalov1998n}. 
It is the same but different format with the area of triangle in 2D space and the volume of tetrahedron in 3D space.
\cite{li2017shape} analysis the generating formula of invariants under different transformations in detail, which can be considered as the Shape-DNA.

There are some researches extend moment invariants into different data formats.  
Some works analyze shape features by extracting invariants for curves in 2D space like \cite{mundy1992geometric,zhao1997affine}. 
For the 3D objects with one channel, researches about 3D affine moment invariants can be found in \cite{lo19893,novotni20033d,xu2008geometric}. 
The invariant for n-D space has been studied in \cite{mamistvalov1998n}.

Color information is important for the recognition of objects. 
Some color descriptors are simple and useful for applications, like color histogram in \cite{funt1995color}, and color moment in \cite{stricker1995similarity}. But they do not consider the color degradation model in reality. Some other color descriptors consider the degradation model of illumination, like works in \cite{gevers1999color,li2009illumination,gong2013moment}.
To consider shape and color degradation in the same framework, there are lots of works. \cite{mindru2004moment} proposed a kind of color moment invariant with Lie group method for only color diagonal transformation.
\cite{gong2017naturally} proposed moment invariants under dual-affine transformation for 2D color images.  Some other invariants for color images can be found in \cite{alferez1999geometric,mindru2004moment,suk2009affine,mo2017shape}.
Affine transformation is considered as the best model to describe the outdoor color degradation statistically in \cite{geusebroek2001color}. 
In this paper, we will present a general framework for the derivation of moment invariants under \textit{dual-affine transformation} for $M$-D objects with $N$ channels. And with 3D color point cloud as an example, we get a set of independent and usable DAMI with $M=3, N=3$.

\section{Dual-Affine Moment Invariants (DAMI)}	
\begin{figure}[t]
	\centering
	\includegraphics[width=0.85\linewidth]{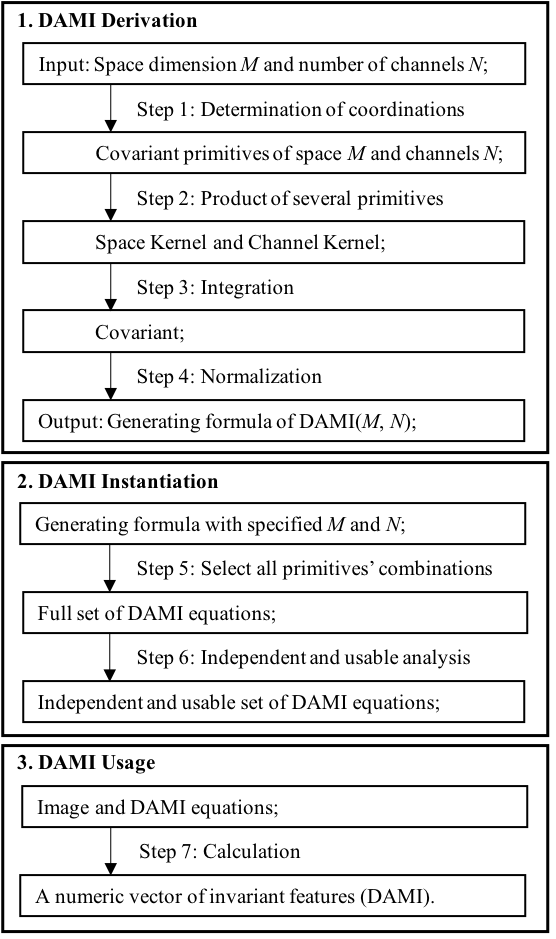}
	\caption{The flowchart of DAMI derivation, instantiation and usage. }
	\label{fig:flow}
\end{figure}

The flowchart of the derivation, instantiation, and usage of DAMI is depicted in Figure \ref{fig:flow}. We will detail the DAMI derivation framework in this section and leave the instantiation and usage part in Section 4 and 5.
The general derivation framework for moment invariants is to first find the \textit{covariant primitive} under a given transformation, and then derive the invariants in the global domain through the integral framework and the normalization factor. Firstly, we give the definition of \textit{covariant primitive} (\textbf{Step 1}) in Section 3.1 and corresponding $SpaceKernel$ and $ChannelKernel$ under $M$-D affine transformation (\textbf{Step 2}) in Section 3.2.
With the proposed $SpaceKernel$ and $ChannelKernel$, we can derive the moment covariants under DAT (\textbf{Step 3}) through integration in Section 3.3. In Section 3.4, we get the generating formula of DAMI after applied normalization (\textbf{Step 4}). Finally, we analysis the null space and a special case of DAMI In Section 3.5 and 3.6.

\subsection{Covariant primitive}
\begin{theorem}		
	The matrix determinant of the coordinates of $M$ points in $M$-D space is an affine covariant primitive.
\end{theorem}

The determinant of coordinates of $M$ points in $M$-D space,  $(x_1,y_1,z_1, (s4)_1,..., (sM)_1)^T$,  $(x_2,y_2,z_2,(s4)_2,...,(sM)_2)^T$, ..., $(x_M,y_M,z_M,(s4)_M,...,(sM)_M)^T$,  is the covariant primitive $P_S(M)$ in shape space as Equation (\ref{equ:sc}):
\begin{equation}\label{equ:sc}
P_{S}(M) = \det(MS(M)) = \left| 
\begin{array}{cccc}
x_1 & x_2 & \cdots & x_M \\
y_1 & y_2 & \cdots & y_M \\
z_1 & z_2 & \cdots & z_M \\
(s4)_1 & (s4)_2 & \cdots & (s4)_M \\
\vdots & \vdots & \ddots & \vdots \\
(sM)_1 & (sM)_2 & \cdots & (sM)_M \\
\end{array}
\right|
\end{equation}	
After $M$-D affine transformation $ATM(M)$, the transformed  $P_{S}^*(M)$ will be Equation (\ref{equ:tsc}):	

\begin{align}\label{equ:tsc}
P_S^*(M) &= \left|MS(M)^* \right| = \left|ATM(M) \cdot MS(M) \right| \notag \\ 
&= \left|ATM(M)\right| \cdot \left|MS(M) \right| = \left|ATM(M)\right| \cdot P_S(M)
\end{align}
where $MS(M)^*$ is the matrix of the transformed coordinates as Equation (\ref{equ:ms}):
\begin{equation}\label{equ:ms}
MS(M)^* = \left(
\begin{array}{cccc}
x_1^* & x_2^* & \cdots & x_M^* \\
y_1^* & y_2^* & \cdots & y_M^* \\
z_1^* & z_2^* & \cdots & z_M^* \\
(s4)_1^* & (s4)_2^* & \cdots & (s4)_M^* \\
\vdots & \vdots & \ddots & \vdots \\
(sM)_1^* & (sM)_2^* & \cdots & (sM)_M^* \\
\end{array}
\right)
\end{equation}
From the result we can see that the determinant before and after the transformation only differs by one scalar coefficient, which is the determinant of the transformation matrix $ATM(M)$. $P_S(M)$ is a covariant primitive in $M$-D space. \textbf{Theorem 1} is proved.

And similarly,  in $N$-D channel space, the matrix determinant of $N$ points with $N$ channels, $(r_1,g_1,b_1,(c4)_1,...,(cN)_1)^T$, $(r_2,g_2,b_2,(c4)_2,...,(cN)_2)^T$, ..., $(r_N,g_N,b_N,(c4)_N,...,(cN)_N)^T$, is covariant primitive in channel space as Equation (\ref{equ:cc}):
\begin{equation}\label{equ:cc}
P_{C}(N)=\det(MC(N))
\end{equation}
where $MC(N)$ is the matrix of the channels' values of $N$ points.
And after $N$-D  affine transformation, the transformed $P^*_{C}(N)$ will be Equation (\ref{equ:tcc}):	
\begin{align}\label{equ:tcc}
P_C^*(N) &= \left|MC(N)^* \right| = \left|ATM(N) \cdot MC(N) \right| \notag \\ 
&= \left|ATM(N)\right| \cdot \left|MC(N) \right| = \left|ATM(N)\right| \cdot P_C(N)
\end{align}

From the results, we can see that the matrix determinant $P_C(N)$ is covariant under affine transformation $ATM(N)$ in $N$-D channel space.

\subsection{Space Kernel and Channel Kernel}

Further, the continued product of several these covariant primitives  is also covariant, as shown in Equation (\ref{equ:kernel}, \ref{equ:kernel2}).

\begin{align} \label{equ:kernel}
SpaceKernel(M,P,O_S) = \prod_{i} P_S(M_i)^{t_i} \\
ChannelKernel(N,Q,O_C) = \prod_{j} P_C(N_j)^{t_j} \label{equ:kernel2}
\end{align}	
In which, $M_i$ and $N_j$ represent  $M$ or $N$ different points and different $i$ and $j$ represent different combinations of points. All points involved in building $ SpaceKernel$ constitute the Point Set $ PS_S $. And all  points involved in building $ ChannelKernel$ constitute the Point Set $ PS_C $. 
The number of points in the point set,  $P = \left|PS_S\right|$ and $ Q = \left|PS_C\right| $, are called the \textbf{degree}.
And $t_i$ and $t_j$ are the order of corresponding primitives separately.
And the $O_S = \sum t_{i}$ and $O_C = \sum t_{j} $ are called the \textbf{order}.

\subsection{Covariant through integration}

To define the $Covariant$ on object $O_{M,N}$, we make an integration of the $SpaceKernel$ and $ChannelKernel$ as Equation (\ref{equ:covariant}):
\begin{align} \label{equ:covariant}
& Covariant(M,N,P,Q,O_S,O_C)(O) \notag \\
&\quad = \iint...\iint SpaceKernel(M,P,O_S) \cdot \notag \\ 
&\quad\quad\quad\quad\quad\quad\quad\quad ChannelKernel(N,Q,O_C) \cdot \notag \\
&\quad\quad\quad\quad\quad\quad\quad\quad dx_1dy_1dz_1 .... ... dx_Ldy_Ldz_L... 
\end{align} 
where $L=|PS_S \cup PS_C|$. 

The relationship between $Covarinat$ before and after $M$-D shape affine and $N$-D channel affine transformation is in Equation (\ref{equ:covariantt}):
\begin{align}\label{equ:covariantt}
&Covariant^*(M,N,P,Q,O_S,O_C)(O^*) \notag \\ 
&\quad = \iint...\iint \prod_i P^*_S(M_i)^{t_i}  \cdot \prod_j P^*_C(N_j)^{t_j} \cdot \notag \\
&\quad\quad\quad\quad\quad\quad\quad\quad dx_1^*dy_1^*dz_1^* .... ... dx_L^*dy_L^*dz_L^*... \notag \\
&\quad = \left| ATM(M) \right|^{O_S+L} \cdot \left| ATM(N) \right|^{O_C} \cdot \notag \\
&\qquad \iint...\iint \prod_i P_S(M_i)^{t_i}  \cdot \prod_j P_C(N_j)^{t_j} \cdot \notag \\
&\quad\quad\quad\quad\quad\quad\quad\quad dx_1dy_1dz_1 .... ... dx_Ldy_Ldz_L...  \notag \\
&\quad = \left| ATM(M) \right|^{O_S+L} \cdot \left| ATM(N) \right|^{O_C} \cdot \notag \\ 
&\qquad\qquad Covariant(M,N,P,Q,O_S,O_C)(O)
\end{align}
We can see that the difference between $Covariant$ and $Covariant^*$ is only one scalar coefficient, which dependents on the transformation matrix $ATM(M)$ and $ATM(N)$. The $Covariant$ is covariant under DAT.

\subsection{Invariant through normalization}

To achieve $Invariant$ on object $O_{M,N}$, we should eliminate the covariant scalar of the $Covariant$. We use a $NormalizationFactor$ consists of some special $Covariant$ to divide the $Covariant$ in Equation (\ref{equ:covariant}). 
	Then we will get $Invariant$ under DAT in $M$-D space with $N$ channels as Equation (\ref{equ:invariant}):
	\begin{align}\label{equ:invariant}
	&Invariant(M,N,P,Q,O_S,O_C)(O) \notag \\
	&\qquad = \frac{
		Covariant(M,N,P,Q,O_S,O_C)
	}{
		NormalizationFactor
	} \notag \\
	&\qquad = \frac{Covariant(M,N,P,Q,O_S,O_C)}{
		\left\{
		\splitdfrac{Covariant(M,N,1,0,0,0)^{O_S+L-N\cdot O_C/2} \cdot}{ Covariant(M,N,1,N,0,2)^{O_C/2}}
		\right\}}
	\end{align}	
	The \textbf{degree} is $L=|PS_S \cup PS_C|$, and the \textbf{order} is $K=O_S+O_C$.
	The $NormalizationFactor$ can be any other one that can eliminate the covariant scalar in the numerator. 	

This definition can be proved in Equation (\ref{equ:invariantt}):

\begin{align}\label{equ:invariantt}
&Invariant^*(M,N,P,Q,O_S,O_C)(O^*) \notag \\
& = \frac{
	Covariant^*(M,N,P,Q,O_S,O_C)
}{
\left\{
	\splitdfrac{
	Covariant^*(M,N,1,0,0,0)^{O_S+L-N \cdot O_C/2} \cdot}{ Covariant^*(M,N,1,N,0,2)^{O_C/2}}
\right\}
} \notag \\
&=  \frac{ 
	\left\{
	\splitdfrac{
	\left| ATM(M) \right|^{O_S+L} \cdot \left| ATM(N) \right|^{O_C} \cdot }{ Covariant(M,N,P,Q,O_S,O_C)}
	\right\}
}{\left\{
	\splitdfrac{(\left| ATM(M) \right|Covariant(M,N,1,0,0,0))^{O_S+L-N\cdot O_C/2} \cdot}
	{(\left| ATM(M) \right|^{N} \cdot \left| ATM(N) \right|^{2} Covariant(M,N,1,N,0,2))^{O_C/2}}
	\right\} } \notag \\
& = \frac{Covariant(M,N,P,Q,O_S,O_C)}{
\left\{
\splitdfrac{
	Covariant(M,N,1,0,0,0)^{O_S+L-N\cdot O_C/2} \cdot }{ Covariant(M,N,1,N,0,2)^{O_C/2}} \right\}} \notag \\
& = 	Invariant(M,N,P,Q,O_S,O_C)(O)
\end{align}	
The invariants before and after the transformation are completely equal. 
$Invariant(M,N,P,Q,O_S,O_C)$ is invariant under DAT in the $M$-D space with $N$ channels, which can be called Dual-Affine Moment Invariants ($DAMI(M,N)$).

\subsection{Null space of DAMI}

As we can see from the definition of covariant primitive in Equation (\ref{equ:sc}) and (\ref{equ:cc}), when the determinant of all primitives equal to 0, the DAMI will be no definition, which can be called the null space of DAMI. 

When the null space happened in the primitives in shape space, it means matrix of the coordinations is singular. And some axis is linear combination of the other axises. The object will be part of a  low-dimensional hyperplane in the original space, where the DMAI has no definition. For this kind of objects, we can find all the independent axises first, say there are $M'$ , and then derive $DAMI(M', N)$ on these axises. That will make the DAMI work again.

When the null space happened in the primitives in channel space, it will be the same with that in shape space. Some channels are linear dependent on others. We can find all the independent channels, say there are $N'$, and then derive $DAMI(M,N')$ on these independent channels. That will make the DAMI work again.  The literature \cite{kostkova2019null} has discussed some about the null space of color space. This situation is rarely seen in RGB images, but may exist in multi-spectral images.

\subsection{Special case}
When $M=2,N=2$, $DAMI(2,2)$ can be used on 2D vector fields with 2 channels. The $ATM(M)$ will be the coordinate transformation of the vector field and the $ATM(N)$ will be the vector transformation at each point. The DAMI will be as Equation (\ref{equ:ami22}):

\begin{align}\label{equ:ami22}
&DAMI(2,2)=Invariant(2,2,P,Q,O_S,O_C)(O) = \notag \\
&\qquad = \frac{
	Covariant^*(2,2,P,Q,O_S,O_C)
}{
\left\{
\splitdfrac{
	Covariant^*(2,2,1,0,0,0)^{O_S+L-O_C}\cdot}{ Covariant^*(2,2,1,2,0,2)^{O_C/2}}
\right\}
} 
\end{align}

When select the same parameters with that in \cite{kostkova2018affine,j2019affine}, we can get the same invariants as Equation (\ref{equ:case22}).
\begin{align}\label{equ:case22}
&DAMI(2,2)(1) = u_{0101}*u_{1010} - u_{0110}*u_{1001}  \notag \\
&DAMI(2,2)(2) = u_{0301}*u_{3010} - u_{0310}*u_{3001} - 3*u_{1201}*u_{2110} + \notag \\
&\quad 3*u_{1210}*u_{2101} \notag \\
&DAMI(2,2)(3) = u_{0501}*u_{5010} - u_{0510}*u_{5001} - 5*u_{1401}*u_{4110} + \notag \\ 
&\quad 5*u_{1410}*u_{4101} + 10*u_{2301}*u_{3210} - 10*u_{2310}*u_{3201} \notag \\
&DAMI(2,2)(4) = u_{0201}^2*u_{2010}^2 - 2*u_{0201}*u_{0210}*u_{2001}*u_{2010} - \notag \\ 
&\quad 4*u_{0201}*u_{1101}*u_{1110}*u_{2010} + 4*u_{0201}*u_{1110}^2*u_{2001} + \notag \\
&\quad u_{0210}^2*u_{2001}^2 + 4*u_{0210}*u_{1101}^2*u_{2010} - \notag \\
&\quad 4*u_{0210}*u_{1101}*u_{1110}*u_{2001}
\end{align}	


\section{Instantiation of $DAMI(3,3)$}	
In this section, we will use the $DAMI(3,3)$ for 3D color objects as an example to demonstrate how to instantiate DAMI.

\subsection{Instantiation}

\begin{figure}[h]
	\centering
	\subfigure[$L=3$]{
		\includegraphics[width=.25\linewidth]{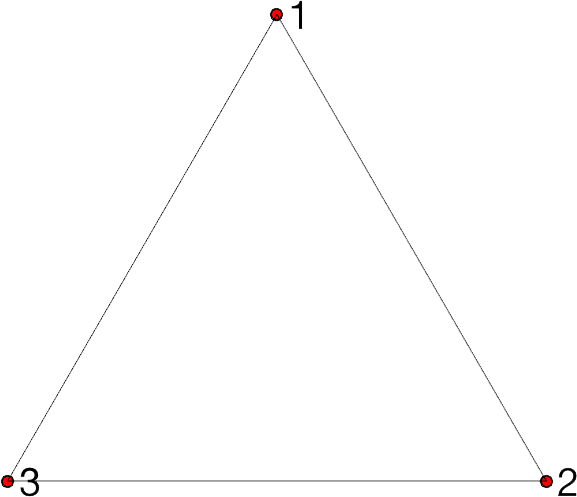}}
	\subfigure[$L=4$]{\label{fig:selectionb33}
		\includegraphics[width=.22\linewidth]{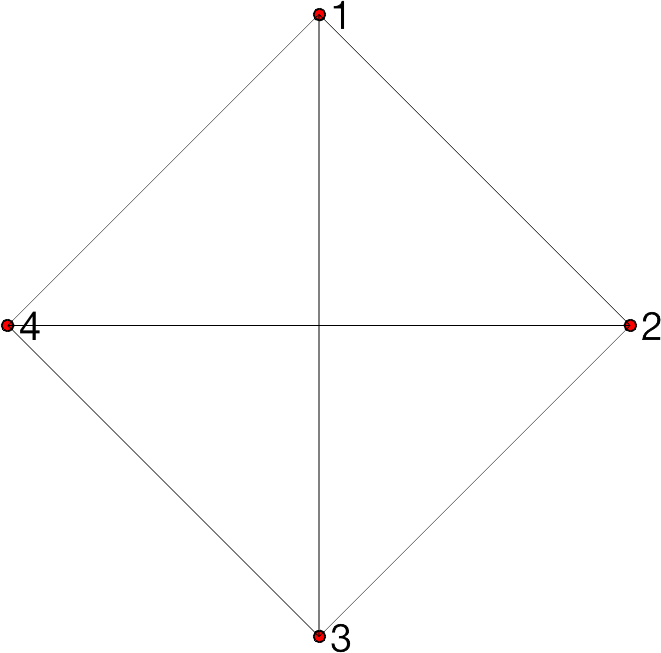}}
	\subfigure[$L=5$]{
		\includegraphics[width=.23\linewidth]{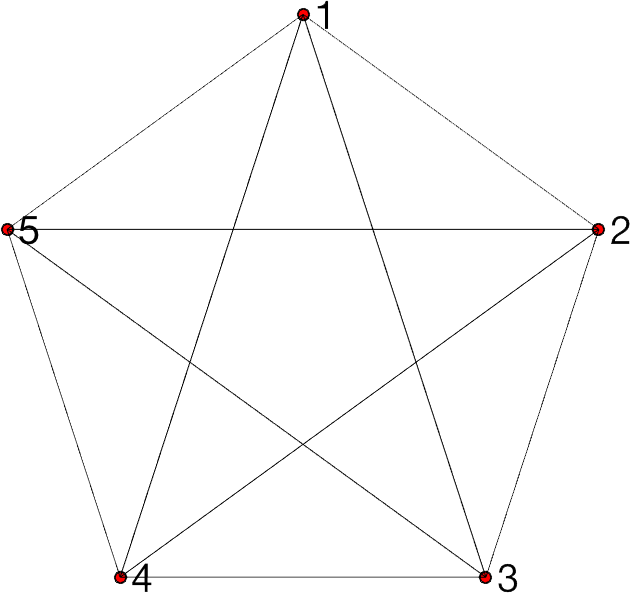}}
	\subfigure[$L=6$]{
		\includegraphics[width=.2\linewidth]{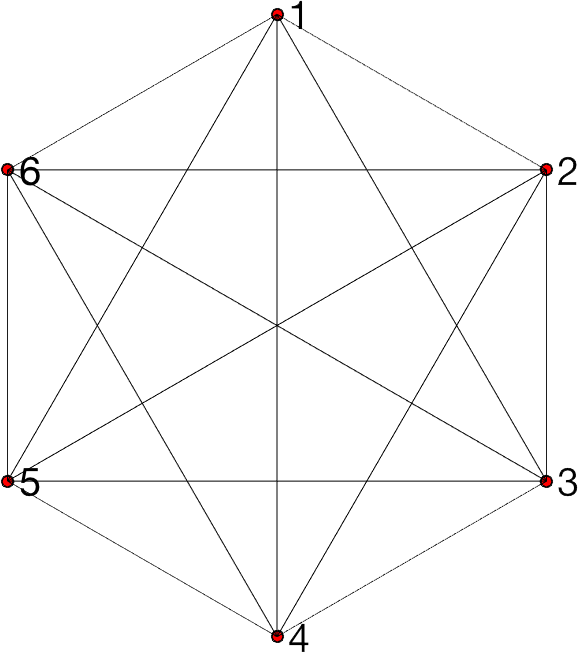}}
	\caption{Visualization of points selection methods with degree $L=3,4,5,6$.}
	\label{fig:selection33}
\end{figure}

To instant the $DAMI(3,3)$, we can select different $SpaceKernel$ and $ChannelKernel$ with different combinations of points. To make the selection clearer and more organized, we use a visualization method, as shown in Figure \ref{fig:selection33}.
In the figure, we use $1,2,3...$ to represent different points' index and each triangle is a \textit{covariant primitive}. $L$ is the degree of the DAMI, which represents how many points has participated in the construction in total. 
Then the $primitives$ constructed with these points can be represented with $D_S(i, j, k)$, $D_C(l, m, n)$,..., and  so on.
Each combination of these triangles (with specified order) would be a $ShapeKernel$ or $ChannelKernel$ of $DAMI(3,3)$. There are limited numbers of triangles in the figure.
And all combinations of $ShapeKernels$ and $ChannelKernels$ in each subfigure make up a Complete Set under the specific degree $L$ and specific order $K$. By the way, Complete Set with lower degree is included in that with higher degree.
What we need to do is just specifying the $i,j,k$ and $l,m,n$ as the different indexes of points.  


Here, we only consider $SpaceKernels$ and $ChannelKernels$ with order and degree both not greater than 4, which are corresponding to Figure \ref{fig:selectionb33}. There are four triangles in total, $(1,2,3)$, $(1,2,4)$, $(1,3,4)$ and $(2,3,4)$. We use them to construct the $covariants$ and specify different orders which not bigger than 4. Then we can get different $DAMI(3,3)$ with different triangle combinations and different parameters $(P,Q,O_S,O_C)$. All of the combinations are listed in Table \ref{tab:com33}. It should be illuminated that some other combinations with other point indexes  does not appear in the Table, because those combinations have the same topological structure with one in the table. One example is $D_S(2,3,4)D_C(2,3,4)$, which does not appear in table, but has the same topological structure with the first element in table, $D_S(1,2,3)  D_C(1,2,3)$. These two will get the same invariant. So just one of them is listed in table. Finally, we get a Complete Set of all combinations with both degree and order not greater than 4. Each one has a specific $ID$ in Table \ref{tab:com33}. And $G$ in the first column of the table is used just to make our enumeration more convenient and orderliness.	
\begin{table}[!t]
	\centering
	\footnotesize
	\caption{All combinations of degree and order not greater than 4, corresponding to Figure \ref{fig:selectionb33}.}
	\label{tab:com33}
	\begin{tabular}{c|c|c|c}
		\toprule
		$G$ & $ID$ &Combination& $(P,Q,O_S,O_C)$ \\
		\midrule
		1&1&$D_S(1,2,3)  D_C(1,2,3)  $  &(3,3,1,1)  \\
		1&2&$D_S(1,2,3)^2D_C(1,2,3)  $  &(3,3,2,1) \\
		1&3&$D_S(1,2,3)  D_C(1,2,3)^2$  &(3,3,1,2)  \\
		1&4&$D_S(1,2,3)^3D_C(1,2,3)  $  &(3,3,3,1) \\
		1&5&$D_S(1,2,3)  D_C(1,2,3)^3$  &(3,3,1,3)  \\
		1&6&$D_S(1,2,3)^2D_C(1,2,3)^2$  &(3,3,2,2)  \\
		\midrule
		\midrule 
		2&7&$D_S(1,2,3)  D_C(1,2,4)  $  &(3,3,1,1)  \\
		2&8&$D_S(1,2,3)^2D_C(1,2,4)  $  &(3,3,2,1)  \\
		2&9&$D_S(1,2,3)  D_C(1,2,4)^2$  &(3,3,1,2)  \\
		2&10&$D_S(1,2,3)^3D_C(1,2,4)  $ &(3,3,3,1)  \\
		2&11&$D_S(1,2,3)  D_C(1,2,4)^3$ &(3,3,1,3)   \\
		2&12&$D_S(1,2,3)^2D_C(1,2,4)^2$ &(3,3,2,2)  \\
		\midrule
		\midrule
		3&13&$D_S(1,2,3)  D_S(1,2,4)D_C(1,2,3)  $ &(4,3,2,1) \\
		3&14&$D_S(1,2,3)^2D_S(1,2,4)D_C(1,2,3)  $ &(4,3,3,1) \\
		3&15&$D_S(1,2,3)D_S(1,2,4)^2D_C(1,2,3)  $ &(4,3,3,1) \\
		3&16&$D_S(1,2,3)  D_S(1,2,4)D_C(1,2,3)^2$ &(4,3,2,2) \\
		3&17&$D_S(1,2,3)  D_C(1,2,3)  D_C(1,2,4)$ &(3,4,1,2) \\
		3&18&$D_S(1,2,3)  D_C(1,2,3)^2D_C(1,2,4)$ &(3,4,1,3) \\
		3&19&$D_S(1,2,3)  D_C(1,2,3)D_C(1,2,4)^2$ &(3,4,1,3) \\
		3&20&$D_S(1,2,3)^2D_C(1,2,3)  D_C(1,2,4)$ &(3,4,2,2) \\
		\midrule
		\midrule
		
		4&21&$D_S(1,2,3)  D_S(1,3,4)D_C(1,2,4)  $ &(4,3,2,1) \\
		4&22&$D_S(1,2,3)^2D_S(1,3,4)D_C(1,2,4)  $ &(4,3,3,1) \\
		4&23&$D_S(1,2,3)  D_S(1,3,4)D_C(1,2,4)^2$ &(4,3,2,2) \\
		4&24&$D_S(1,2,3)  D_C(1,3,4)  D_C(1,2,4)$ &(3,4,1,2) \\
		4&25&$D_S(1,2,3)  D_C(1,3,4)^2D_C(1,2,4)$ &(3,4,1,3) \\
		4&26&$D_S(1,2,3)^2D_C(1,3,4)  D_C(1,2,4)$ &(3,4,2,2) \\
		\midrule
		\midrule
		5&27&$D_S(1,2,3)  D_S(1,2,4)  D_C(1,3,4)  D_C(2,3,4)$ &(4,4,2,2) \\
		5&28&$D_S(1,2,3)  D_S(1,2,4)  D_S(1,3,4)  D_C(2,3,4)$ &(4,4,3,1) \\
		5&29&$D_S(1,2,3)  D_C(1,2,4)  D_C(1,3,4)  D_C(2,3,4)$ &(4,4,1,3) \\
		\bottomrule
	\end{tabular}
\end{table}

\subsection{Calculation}
%
To calculate the $DAMI(3,3)$ in practice, we should expand the $DAMI(3,3)$ to the form of \textit{Moment}.
The basic geometric moment in 2D shape space is listed in Equation (\ref{equ:moment}).	
\begin{equation} \label{equ:moment}
M_{pq}=\iint x^py^q f(x,y) dxdy
\end{equation}
And the central moment is replacing the origin with the centroid, as Equation (\ref{equ:mu}):
\begin{equation}\label{equ:mu}
U_{pq}=\iint (x-\overline{x})^p (y-\overline{y})^q f(x,y) dxdy
\end{equation}
where $\overline{x}$ and $\overline{y}$ represent the center of the image in corresponding axis.

In 3D shape-color space, we define the $M(3,3)$ as Equation (\ref{equ:scm}):
\begin{equation}\label{equ:scm}
M(3,3)_{pqruvw}= \iiint x^py^qz^r R(x,y,z)^u G(x,y,z)^v B(x,y,z)^w dxdydz 
\end{equation}
And the central moment is listed in Equation (\ref{equ:scu}):	
\begin{align}\label{equ:scu}
U(3,3)_{pqruvw}= \iiint &(x-\overline{x})^p(y-\overline{y})^q(z-\overline{z})^r \cdot \notag \\
& (R - \overline{R})^u (G-\overline{G})^v (B-\overline{B})^w dxdydz 
\end{align}
We can expand all the $DAMI(3,3)$ in the form of $U(3,3)$. By the way, some $DAMI(3,3)$ may expand to 0, such as 2,3,8,9,13,17,21,24, in Table \ref{tab:com33}.  Several examples of expanded $DAMI(3,3)$ are listed in Equation (\ref{equ:example1}-\ref{equ:example2}). We use $ID$ listed in Table \ref{tab:com33} to index the 29 $DAMI(3,3)$.
\begin{align}\label{equ:example1}
DA&MI(3,3)(1)\_numerator = \notag \\
&6*u_{001001}* u_{010010}* u_{100100} - 6* u_{001001}* u_{010100}* u_{100010} - \notag \\
&6* u_{001010}*  u_{010001}* u_{100100} + 6* u_{001010}* u_{010100}* u_{100001} + \notag \\
&6* u_{001100}* u_{010001}* u_{100010} - 6* u_{001100}* u_{010010}* u_{100001} \notag \\
DA&MI(3,3)(1)\_denominator = \notag \\
& (- 6* u_{000200}* {u_{000011}}^2 + 12* u_{000011}* u_{000101}* u_{000110} - \notag \\
&6* u_{000020}* {u_{000101}}^2 - 6* u_{000002}* {u_{000110}}^2 + \notag \\
&6* u_{000002}* u_{000020}* u_{000200})^{1/2} * u_{000000}^{5/2} \notag \\
DA&MI(3,3)(1) =\frac{DAMI(3,3)(1)\_numerator}{DAMI(3,3)(1)\_denominator}
\end{align}		
\begin{align}\label{equ:example3}
DA&MI(3,3)(7)\_numerator= \notag \\
&2* u_{000001}* u_{001000}* u_{010010}* u_{100100} - 2* u_{000001}* u_{001000}* \notag \\
&u_{010100}* u_{100010} - 2* u_{000001}* u_{001010}* u_{010000}* u_{100100} + \notag \\
&2* u_{000001}* u_{001010}* u_{010100}* u_{100000} + 2* u_{000001}* u_{001100}* \notag \\
&u_{010000}* u_{100010} - 2* u_{000001}* u_{001100}* u_{010010}* u_{100000} - \notag \\
&2* u_{000010}* u_{001000}* u_{010001}* u_{100100} + 2* u_{000010}* u_{001000}* \notag \\
&u_{010100}* u_{100001} + 2* u_{000010}* u_{001001}* u_{010000}* u_{100100}- \notag \\
&2* u_{000010}* u_{001001}* u_{010100}* u_{100000} - 2* u_{000010}* u_{001100}* \notag \\
&u_{010000}* u_{100001} + 2* u_{000010}* u_{001100}* u_{010001}* u_{100000} + \notag \\
& 2* u_{000100}* u_{001000}* u_{010001}* u_{100010} - 2* u_{000100}* u_{001000}* \notag \\
&u_{010010}* u_{100001} - 2* u_{000100}* u_{001001}* u_{010000}* u_{100010} + \notag \\
& 2* u_{000100}* u_{001001}* u_{010010}* u_{100000} + 2* u_{000100}* u_{001010}* \notag \\
& u_{010000}* u_{100001} - 2* u_{000100}* u_{001010}* u_{010001}* u_{100000} \notag \\
DA&MI(3,3)(7)\_denominator = \notag \\
& (- 6* u_{000200}* {u_{000011}}^2 + 12* u_{000011}* u_{000101}* u_{000110} - \notag \\
&6* u_{000020}* {u_{000101}}^2 - 6* u_{000002}* {u_{000110}}^2 + \notag \\
&6* u_{000002}* u_{000020}* u_{000200})^{1/2} * u_{000000}^{5/2} \notag \\
DA&MI(3,3)(7) =\frac{DAMI(3,3)(7)\_numerator}{DAMI(3,3)(7)\_denominator}
\end{align}		

\begin{equation}\label{equ:example2}
DAMI(3,3)(2,3,8,9,13,17,21,24)=0
\end{equation}		

Considering the length, the others are not listed here.

\section{Experiments for $DAMI(3,3)$}
As listed in Table \ref{tab:com33}, except the 8 DAMIs that are expanded to 0, we get 21 usable DAMIs. In order to give a comprehensive analysis of these DAMIs, we conduct several experiments in this section. 
First, we select a typical object and design an experiment to evaluate the characteristics of DAMI for different types of transformations in Section 5.1. 
Then, we conduct experiments on some special objects with 1 or more symmetric planes in Section 5.2, which may affect the stability of some DAMIs. Through these cases, we can get the characteristics of each DAMI. 
Based on this analysis, we give some guidelines for selecting a high-order invariant kernels. 
Finally, we conduct a classification experiments for the dataset with dual-affine transformations in Section 5.3.

\subsection{Invariance to different transformations}
\begin{figure}[h]
	\centering
	\subfigure[original]{
		\includegraphics[height=0.11\textheight]{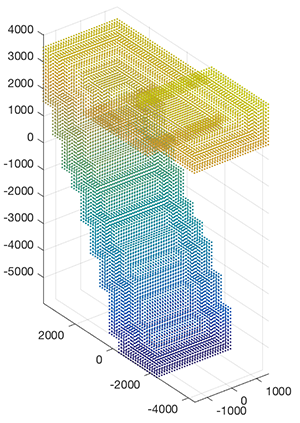}}
	\subfigure[translate]{
		\includegraphics[height=0.11\textheight]{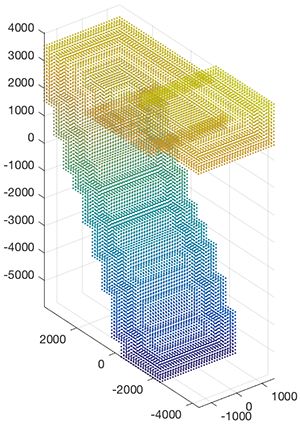}}
	\subfigure[rotation]{
		\includegraphics[height=0.11\textheight]{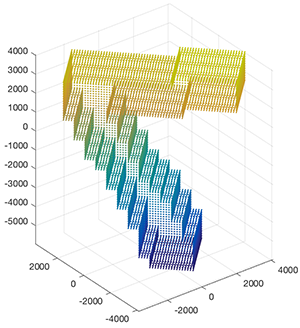}}
	\subfigure[scaling]{
		\includegraphics[height=0.1\textheight]{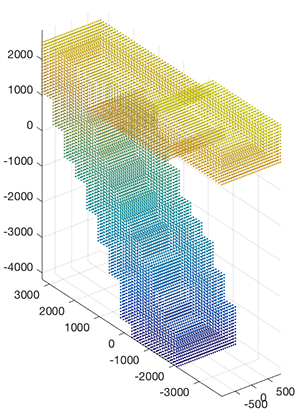}}
	
	\subfigure[affine]{
		\includegraphics[height=0.12\textheight]{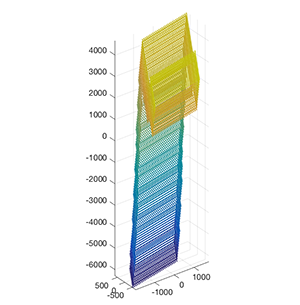}}
	\subfigure[color affine]{
		\includegraphics[height=0.12\textheight]{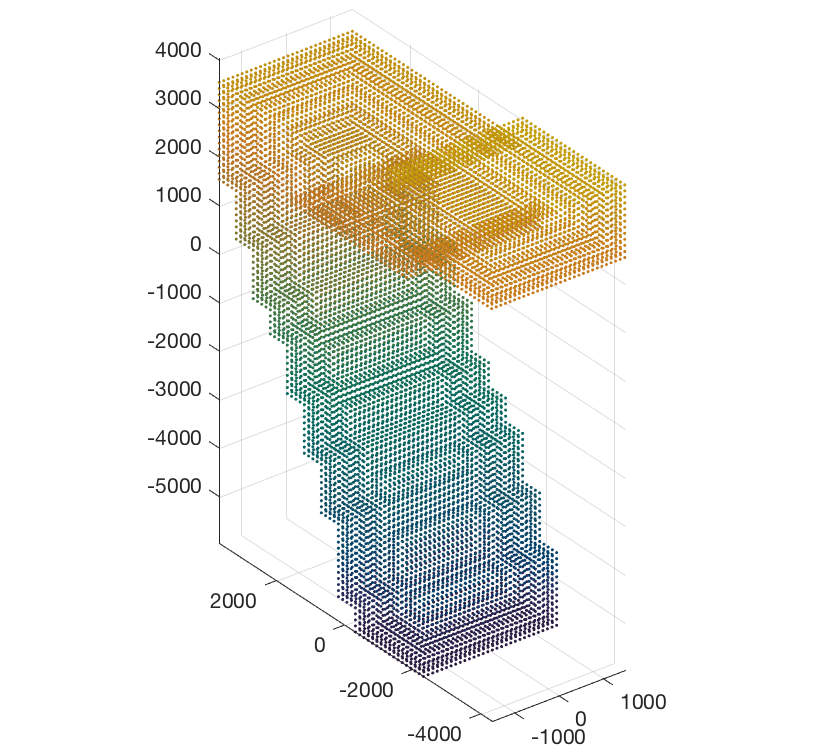}}
	\subfigure[dual-Affine]{
		\includegraphics[height=0.12\textheight]{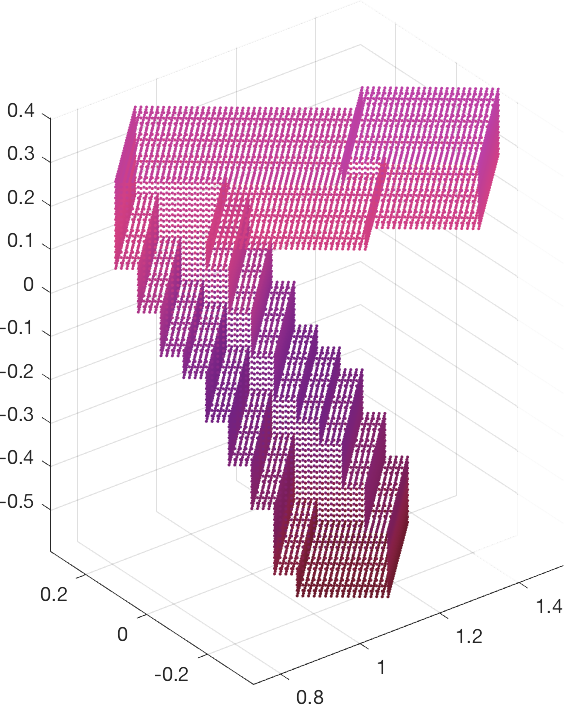}}
	\caption{A 3D color object of point cloud and its  transformed versions.}
	\label{fig:3dmnist}
\end{figure}

To evaluate the invariance of the $DAMI(3,3)$, we use one object from 3D-MNIST dataset\footnote{\url{https://www.kaggle.com/daavoo/3d-mnist}}, which is in 3D points cloud format. Then we assign color for every pixels based the $y$ coordination and generate one 3D color object, as shown in Figure \ref{fig:3dmnist}(a). We transform it with some shape affine transformations, including translation, rotation, scaling, and affine, channel affine transformations and dual-affine transformations. Examples of the 3D color object and its transformed versions are shown in Figure \ref{fig:3dmnist}.

We calculate the 21 $DAMI(3,3)$ for these objects.
To measure the stability of DAMIs, we calculate the coefficient of variation (CV), as Equation (\ref{equ:cv}), for every DAMI to each type of the transformation. The results are listed in Table \ref{tab:cv}. The smaller of the CV is, the more robust of the invariant is. And 0 is the best.

\begin{equation}\label{equ:cv}
c_v = \frac{\sigma}{\mu}
\end{equation}	

\begin{table}[h]
	\tiny
	\centering
	\caption{The CVs of invariants for each type of transformation.}
	\label{tab:cv}
	\begin{tabular}{c|rrrrrrr}
		\toprule
		DAMI 		&   translation    &   rotation   &   scaling   &   affine    &   color-affine    &   DAT & ALL      \\
		\midrule
		1  &  30.4598 &  1.9720 &  68.2011 &  1.9537 &  1.0516 &  2.4390 &   2.9668  \\
		4  &  49.0937 &  1.7006 &   4.1107 &  2.3175 &  0.5987 &  0.9042 &   5.4031  \\
		5  &   0.3933 &  1.5842 &   1.7944 &  2.0706 &  1.8723 &  1.0928 &   3.2553  \\
		6  &   \textbf{0.0000} &  \textbf{0.0000} &   \textbf{0.0000} &  \textbf{0.0000} &  \textbf{0.0130} &  \textbf{0.0130} &   \textbf{0.0084}  \\
		7  & 647.1033 &  1.2743 &   1.6939 &  1.9432 &  4.7739 &  2.0432 &   3.5687  \\
		10 &   1.0987 &  1.8283 &   2.0222 &  2.5839 &  2.8689 &  0.9317 &   2.9907  \\
		11 &  97.7044 &  1.3899 &   1.5922 &  1.9368 &  0.0407 &  0.9426 &   3.3357  \\
		12 &   \textbf{0.0000} &  \textbf{0.0000} &   \textbf{0.0000} &  \textbf{0.0000} &  \textbf{0.0082} &  \textbf{0.0082} &   \textbf{0.0052}  \\
		14 & 525.2069 &  1.2780 &   1.7163 &  0.7668 &  0.0169 &  0.9621 &   1.5313  \\
		15 &   8.1589 &  1.3549 &   2.0697 &  2.5449 &  0.6408 &  0.8617 &  16.5947  \\
		16 &   \textbf{0.0000} &  \textbf{0.0000} &   \textbf{0.0000} &  \textbf{0.0000} &  \textbf{0.0765} &  \textbf{0.0765} &   \textbf{0.0550}  \\
		18 &   1.0154 &  2.4479 &   1.3882 &  2.0443 &  1.1687 &  1.1411 &   3.8883  \\
		19 &   0.3031 &  1.6126 &   1.8729 &  2.0763 &  1.8352 &  1.0757 &   3.0969  \\
		20 &   \textbf{0.0000} & \textbf{0.0000} &   \textbf{0.0000} &  \textbf{0.0000} &  1.5923 &  1.5923 &   4.8507  \\
		22 &   2.3639 &  1.8489 &   0.8868 &  2.5441 &  2.4992 &  1.3321 &   2.8626  \\
		23 &   \textbf{0.0000} &  \textbf{0.0000} &   \textbf{0.0000} &  \textbf{0.0000} &  \textbf{0.1781} &  \textbf{0.1781} &   \textbf{0.1076}  \\
		25 &   0.2547 &  1.7778 &  12.4229 &  2.0604 &  1.5818 &  2.1393 &   5.5918  \\
		26 &   \textbf{0.0000} &  \textbf{0.0000} &   \textbf{0.0000} &  \textbf{0.0000} &  \textbf{0.0079} &  \textbf{0.0079} &   \textbf{0.0064}  \\
		27 &   \textbf{0.0000} &  \textbf{0.0000} &   \textbf{0.0000} &  \textbf{0.0000} &  \textbf{0.0516} &  \textbf{0.0516} &   \textbf{0.0316} \\
		28 &   3.3809 &  1.8488 &   0.5721 &  1.1201 &  1.5067 &  0.8754 &   5.1531  \\
		29 &   1.5408 &  1.2632 &   1.2428 &  2.1340 &  5.3576 &  1.2356 &   3.0848  \\
		\bottomrule			
	\end{tabular}
\end{table}

From the results, we can see that not all $DAMI(3,3)$ are robust to all transformations. In general, invariants with $cv$ less than 1 are considered unstable. We can see that for all transformations of the object, there are 6 DAMIs robust, 6, 12, 16, 23, 26, 27. And No. 20 DAMI is only robust to shape affine transformations.
We will conduct an in-depth analysis of the causes of the errors through experiments in the next subsection.

\subsection{Guidelines for kernel selection}
We analyzed the relationship between the $ShapeKernel$ and symmetric planes of the objects in shape space. In color space, the relationship is the same. 
However, the symmetry in color space rarely occurs on objects.
Therefore, some invariants, with odd-order for some color kernels, are still robust to most objects.	
Another phenomenon we can see from Table \ref{tab:cv} is that the 20th DAMI is not robust to color affine invariants. Its $ColorKernel$ is characterized by the fact that points (1, 2, 3) participate in both the shape $primitive$ and the color $primitive$ construction, and the sum of their orders is odd.


From the above experiments, we can conclude that the calculation of the central moment produces a tiny error for the central moment of all orders due to the tiny error in the calculation of the center of the object.
When the object's symmetric plane is exactly the coordinate plane, the odd moments will be symmetry too and the values of the moments will be close to 0. The absolute values of the error and the moment are on the same order of magnitude and this error will be manifested.
In other cases, the magnitude of the error and the absolute value of the central moment differ greatly in magnitude, the error will be recessive, and the value of the central moment will not be affected.
For some DAMIs, although the expression is not 0, the sum of the addend of the invariant is also close to 0, which will also causes this error to be manifested.
In fact, the necessary and sufficient condition for the invariant to be stable for all objects is that when the invariant is finally expanded, there is one addend that all the central moments in it are even-order. 

Considering the complexity of the actual application, we summarize the DAMI that can be used in different objects.
First, when the objects in the dataset has no more than one symmetric plane, the available invariants are (6, 12, 16, 23, 26, 27). Second, when the objects in the dataset has three or more symmetrical planes, the available DAMIs are (6, 12, 26). Third, for very few cases where colors of the object are also symmetrical in color space, 26 may be unstable, and the available invariants are (6, 12). Fourth, for the case of objects with a single color, the numerator and denominator of DAMI are all 0. The DAMIs are not available. We can use the AMI only in shape space instead.

In order to be able to guide the derivation of higher-order stable invariants, we present some relationship between  invariant kernels and invariant stability: First, A sufficient condition for the stability of DAMI is that the order of each shape kernel and color kernel is even. Second, a necessary condition for DAMI stability is that each point participating in the shape kernel and color kernel has an even number of participations. With these two guideline, we can derive DAMI in high-oder more easily.

\subsection{Classification Experiment}

\begin{figure}[h]
	\centering
	\subfigure[0]{
		\includegraphics[height=.1\textheight]{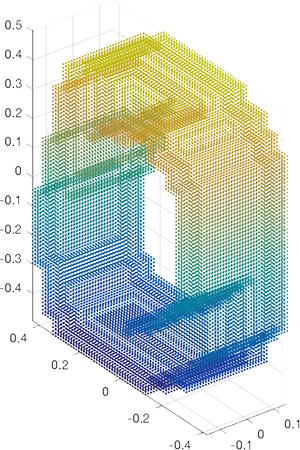}}
	\subfigure[1]{
		\includegraphics[height=.1\textheight]{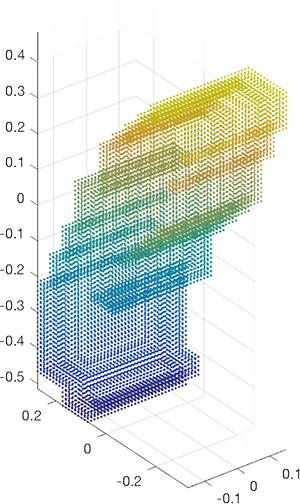}}
	\subfigure[2]{
		\includegraphics[height=.1\textheight]{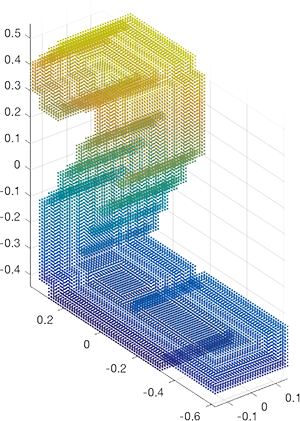}}
	\subfigure[3]{
		\includegraphics[height=.1\textheight]{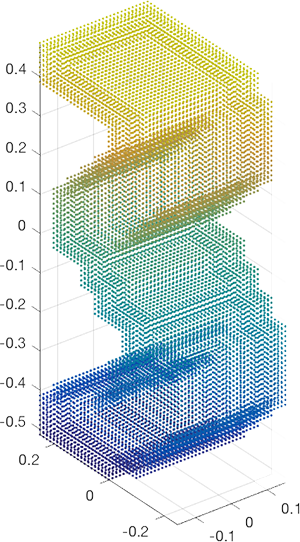}}
	\subfigure[4]{
		\includegraphics[height=.1\textheight]{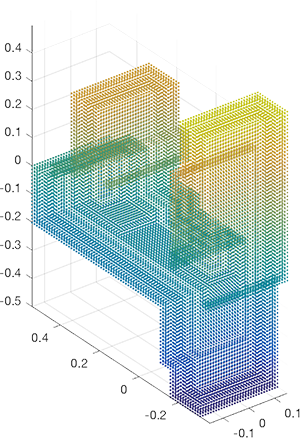}}
	
	\subfigure[5]{
		\includegraphics[height=0.09\textheight]{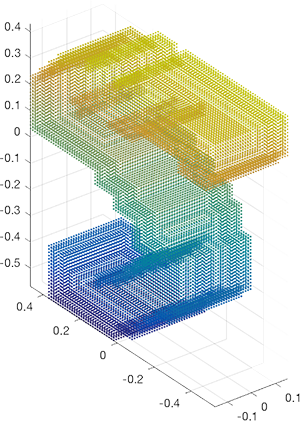}}
	\subfigure[6]{
		\includegraphics[height=0.09\textheight]{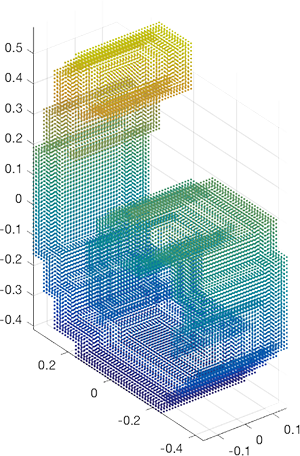}}
	\subfigure[7]{
		\includegraphics[height=0.09\textheight]{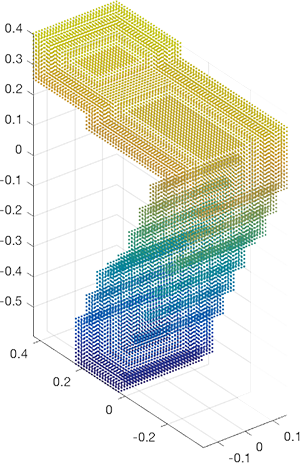}}
	\subfigure[8]{
		\includegraphics[height=0.09\textheight]{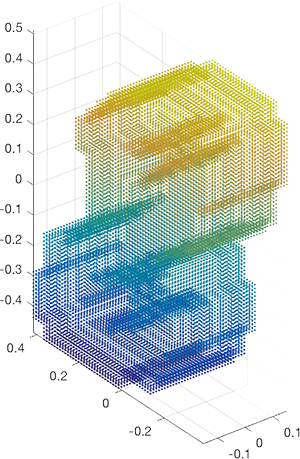}}
	\subfigure[9]{
		\includegraphics[height=0.09\textheight]{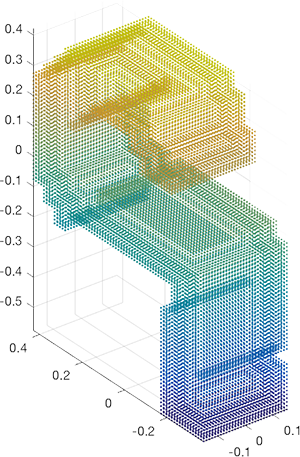}}
	\caption{10 objects from 3D-MNIST dataset.}
	\label{fig:class}
\end{figure}

In order to evaluate the discriminability of invariants, we conduct classification experiments. We select 10 objects from 3D-MNIST dataset, which are shown in Figure \ref{fig:class}. 
Then we conduct 13 DAT on each of them. Finally, we get a dataset with 140 objects of 10 classes.
With the 6 $DAMI(3,3)$, (6, 12, 16, 23, 26, 27), we conduct 10-folds cross validation on the dataset with \textit{K-NearestNeighbor} (KNN). The average accuracy is 100\%. We can conclude that these six invariants have good discriminability.

\section{Conclusion}

In this paper, we propose a general framework to derive Moment Invariants under Dual-Affine Transformation for objects in $M$-dimensional space with $N$ channels. With this framework, we can not only  unify the invariants of objects with different dimensions into one framework, but also get the DAMI of objects with more complex formats. 
This framework integrate the DAMI of different data format into one framework, such as color image in 2D space with 3 channels, 2D vector field in 2D space with 2 channel, 3D color object in 3D shape with 3 channels and objects in higher dimensional space with multi-channels. 
This framework is consist with SCAMI in color images in \cite{gong2017naturally}, and 2D vector field \cite{kostkova2018affine}.
We use the 3D color objects as an example and give out a complete set of $DAMI(3,3)$ with both order and degree not greater than 4. 
Then, we analyzed the root cause of the error in the moment invariant. At the same time, the invariant stability and the choice of the core in the construction process are given. This can play a very good role in the construction of higher moments.
Through invariance analysis, we select six robust DAMIs, which are robust for the objects with no more than 1 symmetric plane. And we evaluated the discriminable of them with classification experiments. The results shown that the DAMI are robust for objects under DAT. We can use this framework to derive DAMI in higher-dimensional space with multiple channels.

\section*{Acknowledgments}
This work is supported by National Key R\&D Program of China (No. 2017YFB1002703), 
National Key Basic Research Program of China (No. 2015CB554507), and National Natural Science Foundation of China (No. 61379082). This work is partially supported by China Scholarship Council (CSC).

%
%
%

\bibliographystyle{model2-names}
\bibliography{refs}

%

\end{document}